\documentclass[sigconf]{acmart}

\renewcommand\footnotetextcopyrightpermission[1]{} 

\usepackage{epsfig}
\usepackage{graphicx}
\usepackage{amsmath}
\usepackage{amssymb}

\usepackage{enumitem}
\usepackage{tablefootnote}
\usepackage[linesnumbered,titlenumbered,ruled,vlined,commentsnumbered,noend]{algorithm2e}
\usepackage{stfloats}
\usepackage{makecell}
\usepackage{colortbl}
\usepackage{enumitem}

\usepackage[font=footnotesize,labelfont=bf]{caption}

\usepackage{multirow}
\usepackage{tabularx}
\usepackage{dsfont}
\usepackage{url}
\usepackage[tight]{subfigure}
\usepackage{color}
\usepackage{subfigure}
\usepackage{amsthm}
\usepackage[export]{adjustbox}
\usepackage{comment}

\usepackage[utf8]{inputenc} 
\usepackage[T1]{fontenc}    
\usepackage{url}            
\usepackage{booktabs}       
\usepackage{amsfonts}       
\usepackage{nicefrac}       
\usepackage{microtype}      

\def\argmax{\operatornamewithlimits{arg\,max}}

\definecolor{dg}{rgb}{0,0.694,0.298}
\definecolor{purple}{rgb}{0.4,0.176,0.569}

\usepackage{pifont}
\usepackage{xspace}
\makeatletter
\DeclareRobustCommand\onedot{\futurelet\@let@token\@onedot}
\def\@onedot{\ifx\@let@token.\else.\null\fi\xspace}
\def\eg{\emph{e.g}\onedot} 
\def\ie{\emph{i.e}\onedot} 
 
\def\etc{\emph{etc}\onedot} 
 
\def\etal{\emph{et al}\onedot}
\makeatother

\AtBeginDocument{%
  \providecommand\BibTeX{{%
    \normalfont B\kern-0.5em{\scshape i\kern-0.25em b}\kern-0.8em\TeX}}}

\begin{document}

\title{\emph{Amora}: Black-box Adversarial Morphing Attack} 









\author{Run Wang$^{1}$, \ Felix Juefei-Xu$^{2}$, \ Qing Guo$^{1,\dagger}$, \ Yihao Huang$^{3}$, \ Xiaofei Xie$^{1}$, \ Lei Ma$^{4}$, \ Yang Liu$^{1,5}$}
\thanks{
Run Wang's email: runwang1991@gmail.com\\
$^{\dagger}$ Qing Guo is the corresponding author~({tsingqguo@gmail.com})}
\affiliation{\institution{$^{1}$Nanyang Technological University, Singapore  \ \ $^{2}$Alibaba Group, USA}}
\affiliation{\institution{$^{3}$East China Normal University, China \ \ $^{4}$Kyushu University, Japan}}
\affiliation{\institution{$^{5}$Institute of Computing Innovation, Zhejiang University, China}}

\renewcommand{\shortauthors}{Run Wang et al.}
\renewcommand{\authors}{Run Wang, Felix Juefei-Xu, Qing Guo, Yihao Huang, Xiaofei Xie, Lei Ma, Yang Liu}

\begin{abstract}

Nowadays, digital facial content manipulation has become ubiquitous and realistic with the success of generative adversarial networks (GANs), making face recognition (FR) systems suffer from unprecedented security concerns. In this paper, we investigate and introduce a new type of adversarial attack to evade FR systems by manipulating facial content, called \textbf{\underline{a}dversarial \underline{mor}phing \underline{a}ttack} (a.k.a. Amora). In contrast to adversarial noise attack that perturbs pixel intensity values by adding human-imperceptible noise, our proposed adversarial morphing attack works at the semantic level that perturbs pixels spatially in a coherent manner. To tackle the black-box attack problem, we devise a simple yet effective joint dictionary learning pipeline to obtain a proprietary optical flow field for each attack. Our extensive evaluation on two popular FR systems demonstrates the effectiveness of our adversarial morphing attack at various levels of morphing intensity with smiling facial expression manipulations. Both open-set and closed-set experimental results indicate that a novel black-box adversarial attack based on local deformation is possible, and is vastly different from additive noise attacks. The findings of this work potentially pave a new research direction towards a more thorough understanding and investigation of image-based adversarial attacks and defenses.


\begin{figure}[t]
	\centering
	\includegraphics[width=0.95\columnwidth]{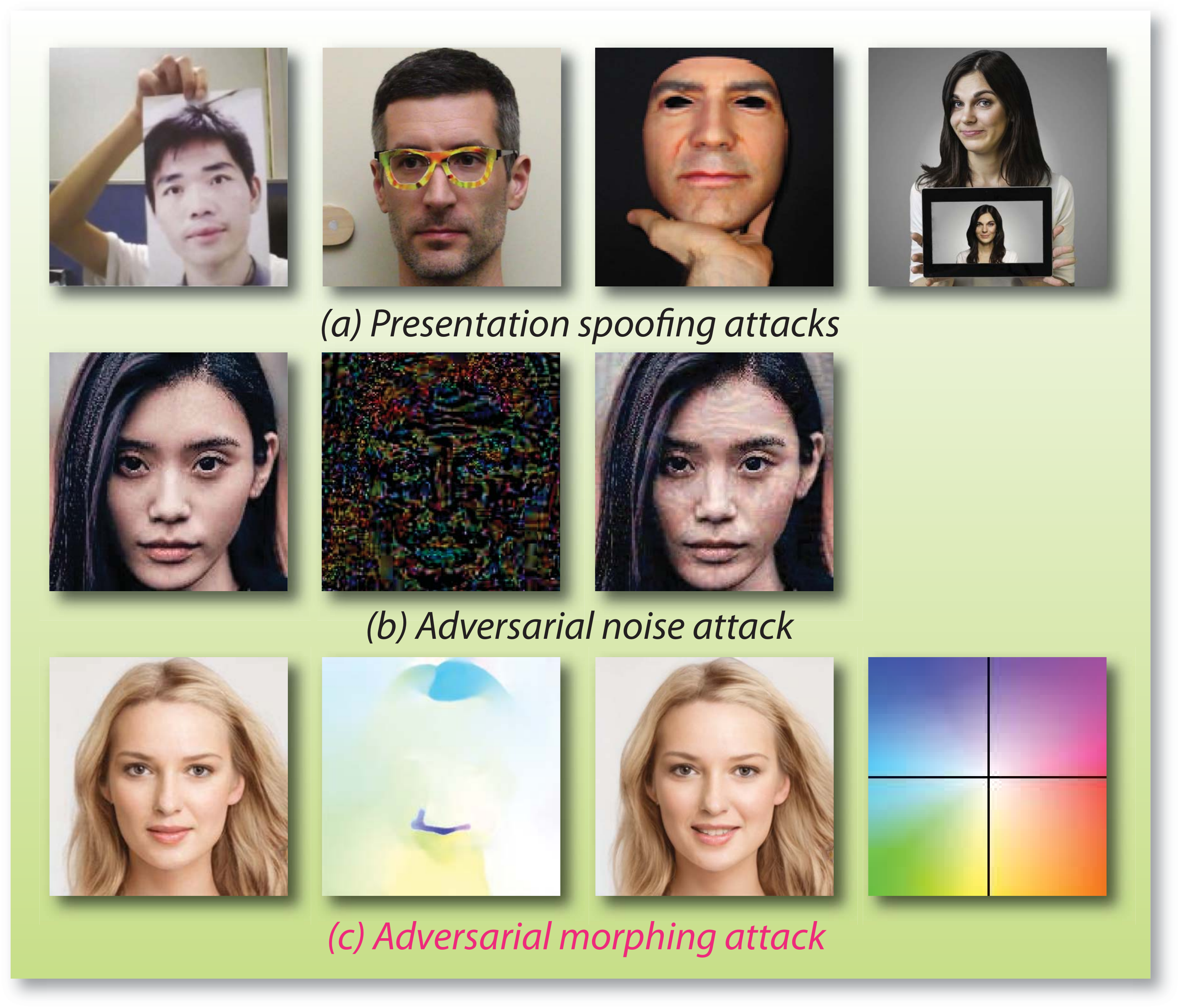}
	\caption{Three typical attacks on the FR systems. (a) presentation spoofing attacks (\eg, print attack \cite{deb2019advfaces}, disguise attack \cite{sharif2016accessorize}, mask attack \cite{maskAttack}, and replay attack \cite{replayAttack}), (b) adversarial noise attack \cite{madry2017towards,deb2019advfaces,goodfellow2014explaining}, (c) proposed \emph{Amora}.}
	\label{Figure:fig1}
\end{figure}
\keywords{Black-box adversarial attack, morphing, face recognition}
\end{abstract}

\ccsdesc[500]{Information systems~Multimedia information systems}
\ccsdesc[500]{Security and privacy~Human and societal aspects of security and privacy}
\ccsdesc[100]{Computing methodologies~Computer vision}

\keywords{Black-box adversarial attack, morphing, face recognition}

\maketitle

\section{Introduction}\label{sec:intro}
Human faces are important biometrics for identity recognition and access control such as security check, mobile payment, \etc. More and more face recognition systems are widely applied in various fields for improving service qualities and securing crucial assets. In recent years, research has shown that current FR systems are vulnerable to various attacks (\eg, presentation spoofing attack \cite{sharif2016accessorize} and adversarial noise attack \cite{biggio2013evasion}), which bring severe concerns to the FR systems deployed in security-sensitive applications (\eg, mobile payment). In this work, we investigate that the FR systems are also vulnerable to another new type of adversarial attack, called adversarial morphing attack, by perturbing pixels spatially in a coherent manner, instead of perturbing pixels by adding imperceptible noise like adversarial noise attack.

Figure~\ref{Figure:fig1} presents three different types of attacks on the FR systems, \ie, presentation spoofing attack, adversarial noise attack, and our adversarial morphing attack. Presentation spoofing attack is a rather simple attack by physically presenting a printed paper, wearing eyeglasses, \etc. In contrast, adversarial noise attack perturbs pixels in images by adding imperceptible noise. Our proposed adversarial morphing attack is another non-additive approach that perturbs pixels spatially in a coherent manner, while adversarial noise attack adds imperceptible noise. Compared with physical spoofing attacks, pixels manipulation by adversarial attack (be it additive noise or Amora) can hardly be detected and poses more severe security concerns than presentation spoofing attack.


The ultimate goal of the black-box adversarial attack (\ie, our focused attack model) is to evade the model by triggering erroneous output and reducing their performance in classification, the basic requirements of which include:

\begin{enumerate}[label=(\arabic*)]
\item First of all, the attack follows \textit{black-box} manner. Generally, attackers cannot gain any knowledge of the model including architecture, parameters, and training data, \etc.
\item The attack should be \textit{transferable}. The crafted examples that attack one target FR system successfully have high success rate on other FR systems.
\item The attack should be \textit{controllable}. Attackers can control the perturbations in generating adversarial faces and query the successful attack faces within limited times.
\item The crafted attack samples should look \textit{visually realistic} and \emph{semantically sound} to humans. It would be highly preferable that the generated adversarial faces do not exhibit any noticeable artifacts .
\end{enumerate}

Our adversarial morphing attack satisfies all of the above requirements. We can effectively attack the FR systems without obtaining any knowledge of the model in a total black-box scenario and the learned attack pattern can be easily transferred to compromising other FR systems. When attacking the FR systems, the adversarial faces are morphed by a learned proprietary morphing field and look visually realistic to the original faces. Figure~\ref{Figure:fig2} illustrates how to morph a face into an adversarial face with the proprietary morphing field to tackle the black-box attack problem. We can also control the intensity of the morphing field to achieve a controllable attack.

Specifically, we first collect numerous frontal and near-frontal faces to attack the FR systems. The successful queries lead to perpetrating faces and flow field pairs, which are used for learning the universal morphing field bases. Then, for any given candidate face during the attack, we assign a proprietary morphing field in each individual attack and morph each and every single face accordingly. The main contributions of our work are summarized as follows:

\begin{figure}[t]
	\centering
	\includegraphics[width=1\columnwidth]{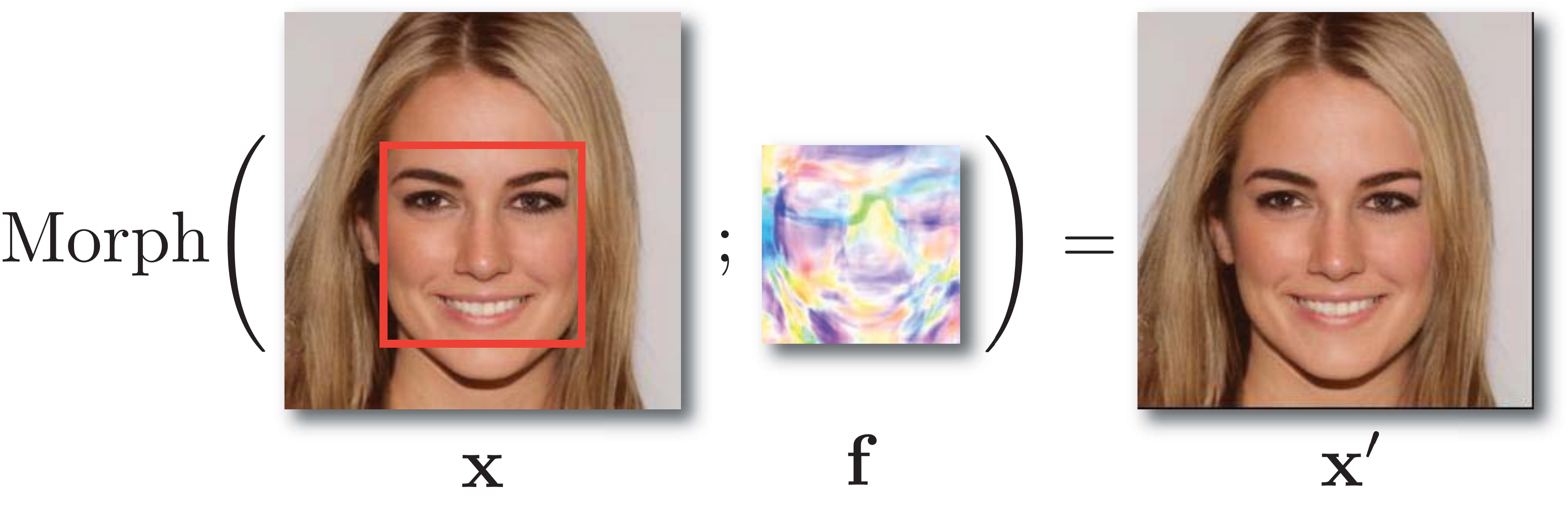}
	\caption{Adversarial morphing attack with proprietary morphing field. The change is very subtle, and $\mathbf{x}$ and $\mathbf{x}$' are indeed different.}
	\label{Figure:fig2}
\end{figure}

\begin{itemize}[leftmargin=*]
\item We introduce a novel type of black-box adversarial attack, namely the black-box adversarial morphing attack (a.k.a. Amora), that morphs facial images with learned proprietary morphing field to generate visually realistic faces that are misclassified by widely adopted FR systems.
\item We devise an effective method based on joint dictionary learning to learn universal morphing field bases and proprietary morphing fields for generating adversarial faces to evade the FR systems.
\item We evaluate the effectiveness of our morphing attack on two popular FR systems (both closed-set and open-set) with smiling facial expression manipulations without obtaining any knowledge of the FR systems. Our learned proprietary morphing fields outperform two competitive baselines in morphing facial images to attack the FR systems.
\item Our research findings hint a new research direction towards semantic-based adversarial attacks and defenses by transforming images in a coherent and natural way, as opposed to adding incoherent noises like adversarial noise attack.
\end{itemize}


\section{Related Work}\label{sec:related}

\textbf{Adversarial Noise Attacks:} The FR systems to be tackled in this work are all deep learning based ones. Studies have demonstrated that deep neural networks are vulnerable to adversarial examples that are widely found in image \cite{goodfellow2014explaining}, texts \cite{ebrahimi2017hotflip}, and audio \cite{zhang2017dolphinattack}, \etc.


\textit{White-box.} White-box adversarial attacks can access the full knowledge of the deep neural networks. A lot of adversarial attack techniques \cite{goodfellow2014explaining,papernot2016limitations,carlini2017towards,moosavi2016deepfool} have been proposed. These techniques could also be applied to attack the FR system. Specifically, the fast gradient sign method (FGSM) \cite{goodfellow2014generative} generates the adversarial examples by performing one step gradient calculation, \ie, adding the sign of gradient of the cost function to the input. Jacobian-based saliency map attack (JSMA) \cite{papernot2016limitations} computes the Jacobian matrix which identifies the impact of the input features on the final output, \ie, which pixel has the most significant influence on the change of the output. C\&W attack \cite{carlini2017towards} is then proposed to generate adversarial attacks by solving the optimization problem whose basic idea of the objective function is to minimize the perturbation such that the output is changed. DeepFool \cite{moosavi2016deepfool} estimates the closest distance between the input and the decision boundary. Based on this, the minimal perturbation is calculated for adversarial examples.

\textit{Black-box.} In a black-box attack setting, the attackers can not access the model's parameters or structure and what they can utilize are only input-output pairs. Current techniques that are applied to generate adversarial samples in a black-box setting mainly rely on transferability, gradient estimation, and heuristic algorithms. Papernot \etal \cite{papernot2017practical} exploit the transferability property of adversarial samples to perform a black-box attack. They trained a substitute model based on the input-output relationships of the original model and crafted adversarial samples for the substituted model in a white-box manner. Narodytska \etal \cite{narodytska2017simple} propose a local-search method to approximate the network gradients, which was then used to select a small fraction of pixels to perturb. Chen \etal \cite{chen2017zoo} utilize the prediction score to estimate the gradients of target model. They applied zeroth-order optimization and stochastic coordinate descent along with various tricks to decrease sample complexity and improve its efficiency. Ilyas \etal \cite{ilyas2018black} adopt natural evolutionary strategies to sample the model's output based on queries around the input and estimate gradient of the model on the input.
In addition, noise-based attacks (white/black) may not be realistic, especially in face recognition domain. Differently, our morphing based method can generate a more realistic face that simulates diverse face transformations.


\textbf{Adversarial Attacks on FR Systems:}
Sharif \etal \cite{sharif2016accessorize} develop a method to fool the FR system, which is realized through printing a pair of eyeglass frames. Different from the noise-based approach, they adopt the optimization to calculate the perturbation on some restricted pixels (on the glasses frames) and they can be modified by a large amount. Similarly, Bose \etal \cite{bose2018adversarial} also generate adversarial attacks by solving the optimization constraints based on a generator network. These techniques are white-box attack, which can be unrealistic in real-world applications. Additionally, some GAN-based attacking techniques have been proposed. Song \etal \cite{song2018attacks} propose a GAN, which adds a conditional variational autoencoder and attention modules, for generating fake faces \cite{wang2019fakespotter,fakelocator}. Deb \etal \cite{deb2019advfaces} propose AdvFaces that learns to generate minimal perturbations in the salient facial regions via GAN. Dong \etal \cite{dong2019efficient} adopt an evolutionary optimization method for generating adversarial samples which is a black-box method. The performance is improved by modeling the local geometry of search directions and reducing the search space. However, they still require many queries. So far there still lacks a work on black-box FR system attack based on pixel morphing.

\textbf{Non-additive Adversarial Attacks:}
The non-additive adversarial attacks start to gain more attention in the research community. \cite{xiao2018spatially} and \cite{alaifari2018adef} are methods that deal with white-box adversarial deformation attacks. In \cite{xiao2018spatially}, the authors use a second order solver (L-BFGS) to find the deformation vector field, while in \cite{alaifari2018adef}, a first-order method is formulated to efficiently solve such an optimization. Our method, in contrast, deals with a black-box setting where we cannot have access to the classifier parameters. Therefore, we need to devise a new method to facilitate such a non-additive adversarial attack. Wasserstein adversarial attack \cite{wong2019wasserstein} is a non-additive attack under the white-box setting that is focused on norm-bounded perturbations based on the Wasserstein distance. The attack covers standard image manipulation such as scaling, rotation, translation, and distortion while our method is able to obtain semantically consistent proprietary morphing field even under a black-box setting. The defense against Wasserstein adversarial attack is proposed \cite{levine2019wasserstein}, which uses optical flow as a way to realize facial image morphing and to carry out black-box attacks on the FR systems. It is worth noting that the proposed attack is not on the optical flow estimation step (see \cite{ranjan2019attacking}), but rather on the face recognition classifier.

\section{Adversarial Morphing Attack}\label{sec:method}
Here, we first briefly review the adversarial noise attack and then present an overview of our adversarial morphing attack. Next, we detail how the proposed adversarial morphing attack method learns the universal morphing field bases and obtains a proprietary morphing field for each individual attack.

\begin{figure}[t]
	\centering
	\includegraphics[width=1\columnwidth]{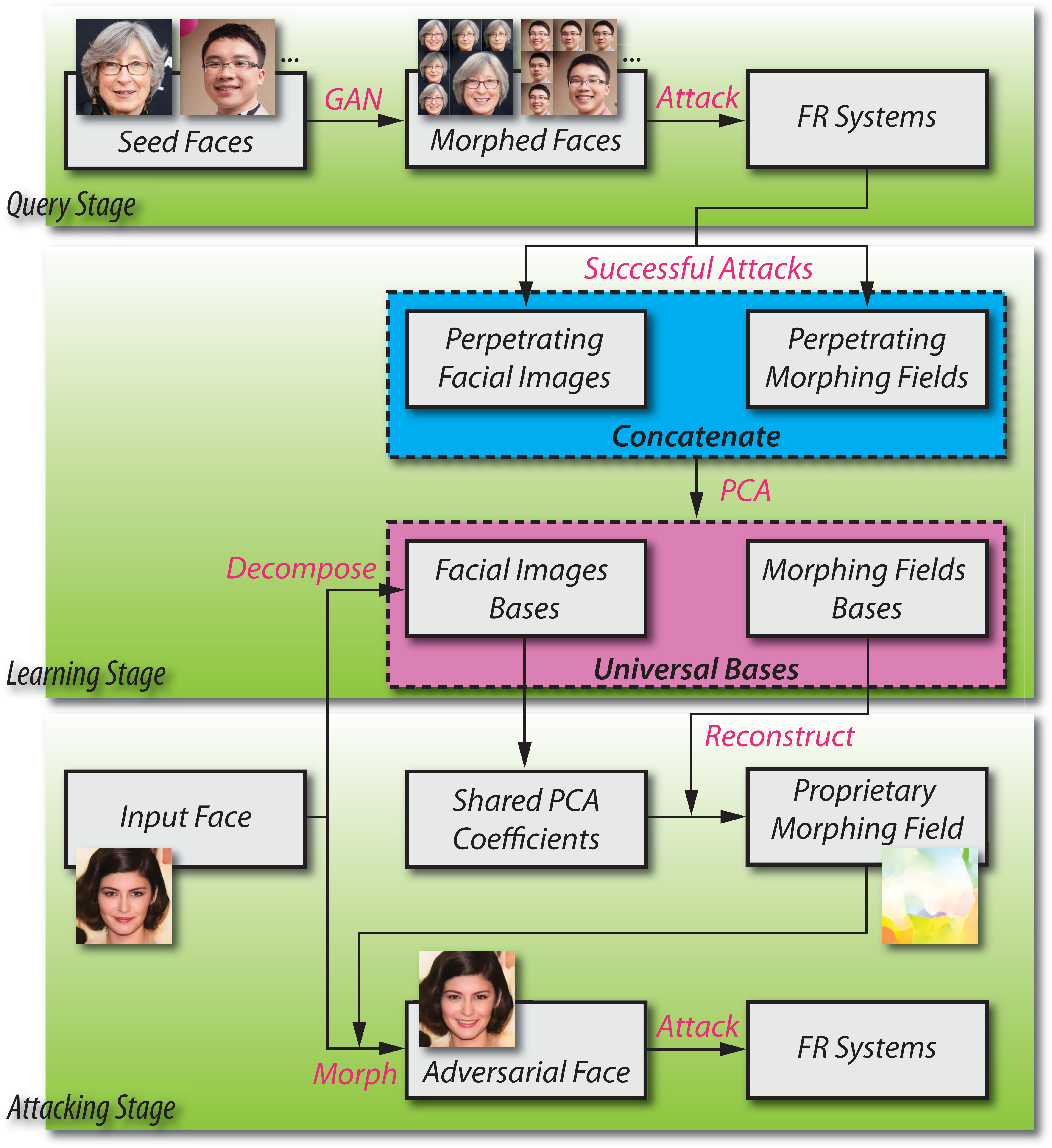}
	 \caption{Overview of the proposed black-box adversarial morphing attack.}
	\label{Figure:fig3}
\end{figure}

\subsection{Brief Review of Adversarial Noise Attack}

In the context of image (gray-scale, RGB, or higher-dimensional) classification problems, let $\mathcal{C}$ be a classifier (shallow or deep) that maps the input image $\mathbf{x}\in \mathds{R}^N$ to a set of discrete and finite categories $\mathcal{L} = \{1,2,\ldots,L\}$. For simplicity, $\mathbf{x}$ is a vectorized single-channel (gray-scale) image with $N$ pixels in total. \emph{Adversarial noise perturbation attack} aims to find a noise or error vector $\mathbf{e}\in\mathds{R}^N$ that is small in $\ell_p$-norm, \ie, imperceptible, such that when added to the input image can cause erroneous classifier output:
\begin{align}
\mathcal{C} (\mathbf{x}+\mathbf{e})  \neq \mathcal{C}(\mathbf{x})~~~~\mathrm{and}~~~~\|\mathbf{e}\|_p ~~\mathrm{is~small}
\end{align}
where $\|\mathbf{e}\|_p = \left(  \sum_{i}^{N}|\mathbf{e}_i|^p  \right)^{1/p}$ for $1\leq p < \infty$ and when $p=\infty$, $\|\mathbf{e}\|_p$ is defined as $\|\mathbf{e}\|_\infty = \max_{i=1,\ldots,N}|\mathbf{e}_i|$. The search for $\mathbf{e}$ under white-box attack is usually done by back-propagating classifier errors all the way to the noise vector $\mathbf{e}$ (see Sec.~\ref{sec:related} for some popular algorithms).
Let $\mathbf{x}'\in \mathds{R}^N$ be a adversarial noise-perturbed counterpart of $\mathbf{x}$, the image modification procedure is summarized as:
\begin{align}
    \mathbf{x}' = \mathrm{Perturb}(\mathbf{x}; \mathbf{e}) = \mathbf{x} + \mathbf{e}
\end{align}
Whereas, in this work, we are seeking a non-additive image modification (spatial perturbation of pixels \textit{c.f.} pixel value perturbation) with the aid of optical flow:
\begin{align}
    \mathbf{x}' = \mathrm{Morph}(\mathbf{x}; \mathbf{f}^h, \mathbf{f}^v)
    \label{eq:ama}
\end{align}
where $\mathbf{f}^h \in \mathds{R}^N$ and $\mathbf{f}^v\in \mathds{R}^N$ are the horizontal and vertical flow fields, respectively. The concatenated whole field is expressed as $\mathbf{f}$. The actual $\mathrm{Morph}(\cdot)$ function works on 2D images, so there is an implicit step to map the vectorized image and flow fields back to 2D. Modifying images according to Eq.~(\ref{eq:ama}) to fool the classifier (with erroneous prediction) is what we call the \emph{adversarial morphing attack}.

\subsection{Overview of Adversarial Morphing Attack}

Under the black-box adversarial morphing attack settings, the attackers do not have access to the model parameters (\ie, deep learning based classifier), and thus obtaining the morphing field uniquely to each attack image by back-propagating the classifier errors through the network parameters is not feasible. In order to obtain a proprietary morphing field, we propose to learn a set of universal morphing field bases, and through which, the proprietary morphing field can be reconstructed for each individual attack image. In the next two subsections, we will detail the learning procedure of the universal morphing field bases as well as how to assign a proprietary morphing field.

Figure~\ref{Figure:fig3} outlines the framework of our adversarial morphing attack to learn universal morphing field bases and obtain a proprietary morphing field for each individual attack image. It contains three essential stages: (1) query stage, (2) learning stage, and (3) attacking stage. In the query stage, we collect a set of seed faces and generate morphed faces with GAN to attack the FR systems. In the learning stage, the successful attacks lead to the collection of perpetrating facial images as well as the morphing fields, and from which we will learn the universal morphing field bases using joint dictionary learning through principal component analysis (PCA) \cite{wold1987principal} by concatenating the perpetrating facial images and corresponding perpetrating morphing fields in the learning framework. Finally, in the attacking stage, we obtain a proprietary morphing field for an input face to morph it adversarially for attacking the face recognition systems.

\subsection{Learning Universal Morphing Field Bases}

In preparing the training images to learn universal morphing field bases, we first collect numerous frontal and near-frontal faces to generate images with consecutive subtle facial expression change. Specifically, we leverage the power of GAN in digital image manipulation to create a large amount of smiling facial images for each seed face. These morphed smiling facial images are smiled in a controllable way while ensuring other parts relatively unchanged. The consecutive set of smiling faces allows us to accurately capture the morphing field of smiling with optical flow that represents the motion of the pixels between two images (see details in Section~\ref{sec:exp}).

Once we obtain a large number of (image, morphing field) pairs: $(\mathbf{x}_i,\mathbf{f}_i)$ whose resulting morphed images $\mathbf{x}'_i$ are successful in attacking the model, we can capitalize on the fact that there exists a correlation between the image $\mathbf{x}_i$ and the `perpetrating' morphing field $\mathbf{f}_i$. There are many ways to learn the 1-to-1 mapping between a given face and its morphing field. Our method draws inspiration from joint dictionary learning for cross-domain generation/matching problems such as super-resolution \cite{yang-superres}(eq.24 therein), hallucinating full face from periocular region \cite{juefei2014hallucinating}(eq.4 therein), and \cite{juefei2015nir,juefei2015encoding,juefei2016fastfood,juefei2016learning,juefei2015pokerface,juefei2016multi,pal2016discriminative,juefei2015single}. In this work, we use PCA, a simple yet effective method, to learn the universal morphing field bases. By stacking the corresponding counterparts as part of the same data point, we are implicitly enforcing a 1-to-1 mapping between the two domains (\ie, image and morphing field). Once such a mapping is established between $\mathbf{x}_i$s and $\mathbf{f}_i$s, we can potentially reconstruct a proprietary `perpetrating' morphing field for any image of interest. The gist here is to `stack' the two dictionaries (PCA basis for face and morph field) during the optimization process so that they can be learned \textbf{jointly}. By doing so, the PCA coefficient vector is `forced' to be shared among the two dictionaries as a `bridge' so that cross-domain reconstruction is made possible. That said, if two sets of PCA bases are learned separately, such projection would not make sense.

The training data matrix $\mathbf{X}\in\mathds{R}^{3N\times M}$ contains concatenated mean-subtracted image and flow field pairs in its columns, with a total of $M$ instances:
\begin{align}
    \mathbf{X} = \begin{bmatrix}
    \boldsymbol{\Lambda}_x(\mathbf{x}_1 - \boldsymbol{\mu}_x), & \ldots, & \boldsymbol{\Lambda}_x(\mathbf{x}_i - \boldsymbol{\mu}_x), & \ldots \\
    \boldsymbol{\Lambda}_h(\mathbf{f}_1^h - \boldsymbol{\mu}_h), & \ldots, & \boldsymbol{\Lambda}_h(\mathbf{f}_i^h - \boldsymbol{\mu}_h), & \ldots \\
    \boldsymbol{\Lambda}_v(\mathbf{f}_1^v - \boldsymbol{\mu}_v), & \ldots, & \boldsymbol{\Lambda}_v(\mathbf{f}_i^v - \boldsymbol{\mu}_v), & \ldots
    \end{bmatrix}
\end{align}
where $\boldsymbol{\Lambda}_x \in\mathds{R}^{N\times N}$, $\boldsymbol{\Lambda}_h \in\mathds{R}^{N\times N}$, and $\boldsymbol{\Lambda}_v \in\mathds{R}^{N\times N}$ are diagonal dimensionality weighting matrices for the image, and the two flow fields, respectively. By setting certain diagonal elements to 0 in $\boldsymbol{\Lambda}_x$, $\boldsymbol{\Lambda}_h$, and $\boldsymbol{\Lambda}_v$, we can arbitrarily select the region of interest (ROI) in the optimization. In this work, the ROI is tight crop on the face region as shown in Figure~\ref{Figure:fig2} to ignore image deformations outside the face region that may contribute to the successful attacks. However, it might be interesting to explore that in a future work.
The bases $\mathbf{w}_i \in\mathds{R}^{3N}$ can be obtained with the following optimization:
\begin{align}
    J(\mathbf{w}) &= \argmax_{\|\mathbf{w}\|=1}{  \mathbf{w}^\top \mathbf{X}\mathbf{X}^\top\mathbf{w} }
    = \argmax_{\|\mathbf{w}\|=1}{ \frac{ \mathbf{w}^\top \mathbf{S}_1 \mathbf{w}}{ \mathbf{w}^\top \mathbf{S}_2 \mathbf{w} }  }
\end{align}
where $\mathbf{S}_1 = \mathbf{X}\mathbf{X}^\top$ is the covariance matrix with $\mathbf{X}$ being mean-subtracted, and $\mathbf{S}_2 = \mathbf{I}$. The objective function translates to a generalized Rayleigh quotient and the maximizer $\mathbf{w}$ can be found by solving the eigen-decomposition of $\mathbf{S}_2^{-1}\mathbf{S}_1$ which is $\mathrm{eig}(\mathbf{S}_1)$.

\subsection{Assigning Proprietary Morphing Field}

For simplicity, let us assume $\boldsymbol{\Lambda}_x=\boldsymbol{\Lambda}_h=\boldsymbol{\Lambda}_v=\mathbf{I}$. The learned universal bases (principal components) can be broken down to an image portion $\mathbf{w}_x\in\mathds{R}^N$, as well as morphing fields portions $\mathbf{w}_h\in\mathds{R}^N$ and $\mathbf{w}_v\in\mathds{R}^N$. When a potential attack image $\mathbf{y}\in\mathds{R}^N$ comes in, we can decompose it with the top-$K, (K<N)$ image portion bases $(\mathbf{W}_x\in\mathds{R}^{N\times K})$ and obtain the PCA projection coefficient vector $\boldsymbol{\alpha}\in\mathds{R}^K$:
\begin{align}
    \boldsymbol{\alpha} = (\mathbf{W}_x^\top \mathbf{W}_x)^{-1} \mathbf{W}_x^\top \mathbf{y}
\end{align}

\begin{figure}[t]
	\centering
	\includegraphics[width=1\columnwidth]{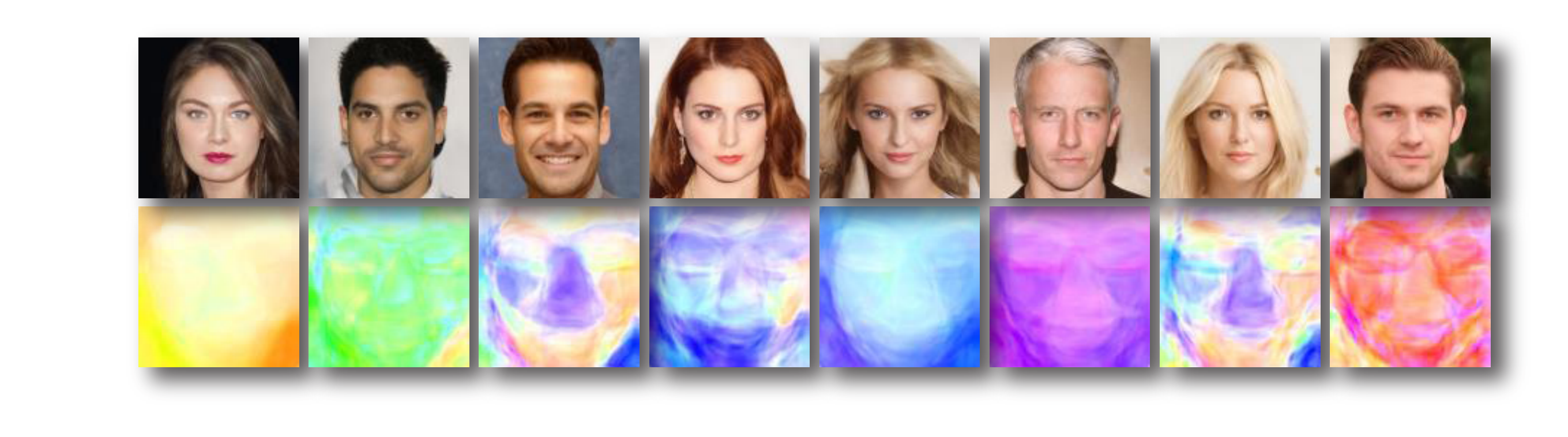}
	 \caption{Examples of the proprietary morphing fields. (T) facial images, (B) the tight cropped faces' corresponding proprietary optical flow.}
	\label{Figure:fig4}
\end{figure}

By forcing consistent PCA representations during training for both image and flow field, the mapping between the two is implicitly learned. Therefore, we can obtain the proprietary flow field $\mathbf{f}_y \in \mathds{R}^{2N}$ by reconstructing using the shared coefficients $\boldsymbol{\alpha}$ and the flow field portion $\mathbf{W}_f = [\mathbf{W}_h;\mathbf{W}_v] \in \mathds{R}^{2N \times K}$ of the bases: $\mathbf{f}_y = \mathbf{W}_f \boldsymbol{\alpha}$.
Examples of proprietary morphing fields are shown in Figure~\ref{Figure:fig4}. The first row is the originally given faces and the second row is their proprietary morphing fields learned by our proposed approach.

\section{Experiments}\label{sec:exp}
In this section, we evaluate the effectiveness of our adversarial morphing attack in evading closed-set and open-set FR systems under total black-box scenario.
We also demonstrate the transferability of our adversarial morphing attack by investigating whether one adversarial face can be transferred to different FR systems. Furthermore, we build several baselines to evaluate whether our learned proprietary morphing field could indeed be effective in evading the FR systems under the `imperceptible' assumption.

\subsection{Experimental Settings}

\textbf{FR systems.} We study two popular FR systems, including VGG-Face \cite{keras-vggface} with VGG16 and ResNet50 as their architectures. In testing, facial images are morphed by the learned proprietary optical flow field. Then, optical flows at different intensity levels are calculated to query the successful attack morphing fields. Finally, we use these morphing fields to evaluate the transferability of our attack.

\noindent\textbf{Dataset.} We conduct experiments on the CelebFaces Attributes (\textit{CelebA}) \cite{liu2015faceattributes} and \textit{CelebA-HQ} \cite{karras2017progressive} datasets. We select 2K and 1K identities from this dataset to train two FR systems VGG-Face (VGG16) and VGG-Face (ResNet50), respectively, for evaluating their robustness against our proposed morphing attack. \textit{CelebA-HQ} \cite{pggan} is a high-quality version of the \textit{CelebA} dataset with more than 30K facial images in 1024x1024 resolution. We use this high quality facial images to generate smiling faces with the latest StyleGAN \cite{karras2019style,abdal2019image2stylegan} to attack the FR systems.

%

%

\noindent\textbf{Evaluation metrics.} In attacking the FR systems, a morphed facial image should be imperceptible to human eyes and is an adversarial face to the FR systems, which leads to misclassification. Thus, some similarity measures between the morphed and the original facial images are needed in order to evaluate the performance of our attack. We report the mean attack success rate as well as the intensity of the morphing fields. Specifically, we employ the \textit{Euclidean} distance ($\ell_2$-norm) and $\ell_\infty$-norm as metrics to measure the intensity of the morphing fields. 

\noindent\textbf{Implementation.} Our numerous smiling faces are morphed by StyleGAN and their optical flows are generated using FlowNet2 \cite{IMKDB17}.
StyleGAN utilizes the extended latent space $W^+$ for editing images with a loss function that is a combination of VGG-16 perceptual loss and pixel-wise MSE loss \cite{stylegan_encoder,abdal2019image2stylegan}, which allows us to generate smiling faces without any visible identity shifts \cite{korshunov2018deepfakes} and to capture subtle smiling variations with optical flows.
%
FlowNet2 \cite{flownet2-pytorch} estimates optical flow between two consecutive images in a learnable way. Here, we use FlowNet2 to generate optical flows of facial images which attack the FR systems successfully.
All the experiments were performed on a server running Ubuntu 16.04 system on an 18-core 2.30 GHz Xeon CPU with 200 GB RAM and an NVIDIA Tesla M40 GPU with 24 GB memory.

\subsection{Experimental Results}

Here, we report the experimental results on the \textit{CelebA} dataset. Specifically, we mainly investigate the following research questions: 1) the effectiveness of the learned proprietary morphing fields in attacking the FR systems, 2) the relation between the attack success rate and the intensity of the morphing fields, 3) the transferabilities of our adversarial morphing attack, 4) the capabilities in attacking open-set face recognition systems, and 5) the performance in comparison with baselines.


\noindent\textbf{Data preparation.} We collect $182$ identities with frontal-face or near frontal-face from \textit{CelebA-HQ} as seed faces to generate morphed smiling faces with StyleGAN. Table~\ref{Table:occ} shows the number of identities and facial images in our training dataset, \textit{CelebA-HQ} and \textit{CelebA}. In attacking the FR systems, we randomly select $120$ identities including more than $1,000$ facial images and morph them with proprietary morphing field to evaluate the robustness of the FR systems against our adversarial morphing attack.
\begin{table}[t]
	\centering
	\small
	\caption{The number of face identities and images of our collected dataset in the query stage.}
	\begin{tabular}{c|cccc}
	\toprule
		 & Seed Face & \textit{CelebA-HQ} & \textit{CelebA}   \\
	\midrule
		Id  &   182      &   6,217    & 10,177    \\ \midrule
		Img &   18,200   &  30,000    & 202,599   \\
	\bottomrule
	\end{tabular}
	\label{Table:occ}
\end{table}
Table~\ref{Table:eval} presents the detailed number of identities and facial images in attacking the two popular FR systems. To obtain successfully attacked perpetrating facial images and the pairing morphing fields, we need to attack some FR systems in the query stage. Table~\ref{Table:eval} presents the number of identities and their corresponding facial images in training the two popular FR systems (\eg, VGG-Face with VGG16 and ResNet50).

\begin{table}[t]
    \centering
    \small
    \caption{The number of face identities and facial images used for attacking and training the 2 popular FR systems.}
    \begin{tabular}{l|c|c|c|c}
    \toprule
    & \multicolumn{2}{c|}{Attacking}          & \multicolumn{2}{c}{Training}  \\
    & Id                 & Img & Id                    & Img \\ \midrule
    \midrule
    \multicolumn{1}{c|}{\begin{tabular}[c]{@{}c@{}}VGG16\end{tabular}}    & 120 & 1,141  & 2,000     & 42,256 \\ \midrule
    \multicolumn{1}{c|}{\begin{tabular}[c]{@{}c@{}}ResNet50\end{tabular}} & 120 & 1,159  & 1,000     & 21,152 \\ \bottomrule
\end{tabular}
\label{Table:eval}
\end{table}


\noindent\textbf{Collecting successful attack morphing fields.} In the query stage, we generate numerous morphed facial images to attack FR systems to obtain perpetrating facial images and perpetrating morphing field pairs for learning the universal morphing field bases. Table \ref{Table:attack} presents the detailed statistical data of the successful attack pairs. More than $153$ and $148$ identities from $182$ seed faces and nearly $10,000$ facial images have successfully attacked the two popular FR systems. In order to obtain large numbers of facial image and morphing field pairs to learn a representative universal morphing field bases, we apply the following strategies to determine a successful attack. 1) causing erroneous output, which directly misclassifies the identity; 2) significant reducing the performance of FR systems, which has low confidence in predicting the identity, \ie, the confidence score is lower than the set threshold $\gamma=0.6$.

\noindent\textbf{Metrics.} We use a series of metrics to explore the relation between the attack success rate and the intensity of proprietary morphing field. Specifically, we employ two popular metrics $\ell_2$-norm and $\ell_\infty$-norm to measure the intensity of proprietary morphing fields.

\noindent\textbf{Adversarial morphing attack.} In morphing facial images with the learned proprietary morphing fields, we need to identify the range of the intensity of morphing fields and control them to investigate the relation between attack success rate. We first obtain the range of the intensity of the morphing fields from the raw morphing fields which have successfully attacked the FR systems in the query stage. The intensity of the morphing fields are measured with $\ell_2$-norm and $\ell_\infty$-norm. To investigate the distribution of the intensity of raw morphing fields, we find that most of the $\ell_2$-norm and $\ell_\infty$-norm value lie in a fixed range. Thus, the intensity of proprietary morphing fields is split into several groups according to the fixed range value to evaluate their effectiveness in attacking the target FR systems.

\noindent\textbf{Assigning proprietary morphing fields.} Table~\ref{Table:l2_to_rate} and Table \ref{Table:l_infinte_to_rate} consolidate the attack success rates vs. different intensity of {proprietary optical flow field} with $\ell_2$-norm and $\ell_\infty$-norm, respectively. The intensity of proprietary morphing fields is mainly split into three groups according to the distribution of the raw morphing fields.
In measuring the intensity of proprietary morphing fields with $\ell_2$-norm, the three groups of intensity are as follows: 1)  $[2,10]$ with a step value $2$; 2) $[100,200]$ with a step value $10$; 3) $[200,600]$ with a step value $100$. In measuring the intensity of the proprietary morphing fields with $\ell_\infty$-norm, the three groups of intensity are as follows: 1) $[0.1,0.5]$ with a step value $0.1$; 2) $[1.0,2.0]$ with a step value $0.1$; 3) $[2.0,6.0]$ with a step value $1.0$.
\begin{table}[t]
	\centering
    \small
	\caption{The number of facial identities and morphing fields that successfully attacked the FR systems in the query stage.}
	\begin{tabular}{c|ccc}
	\toprule
		 & {\begin{tabular}[c]{@{}c@{}}VGG16\end{tabular}} & {\begin{tabular}[c]{@{}c@{}}ResNet50\end{tabular}}\\
	\midrule
		Identity  &   153    &  148      \\  \midrule
		Morphing Field  &   9,621  &  10,536   \\
	\bottomrule
	\end{tabular}
	\label{Table:attack}
	\vspace{-15pt}
\end{table}

Figure~\ref{Figure:fig6} (L) and (C) plot the relation between the attack success rate and the modulated flow field on $\ell_2$-norm and $\ell_\infty$-norm. We can find that the attack success rate increases with the intensity of proprietary morphing field. Since the intensity range spans two orders of magnitude, we present the plots in Figure~\ref{Figure:fig6} (L) and (C) in semi-log on the x-axis. Experimental results show that VGG-Face with VGG16 as backend architecture is more vulnerable than VGG-Face with ResNet50 as the backend architecture. Amora reaches nearly $60\%$ attack success rate in attacking the two popular FR systems, \ie, VGG-Face (VGG16) and VGG-Face (ResNet50). Additionally, we also explore the relation between the attack success rate and the modulated flow field on the multiplier $\delta$ for enhancing the intensity of proprietary morphing fields. The range of the coefficient $\delta$ is from $0.2$ to $2.0$ with a step value $0.2$.

\begin{figure}[t]
\centering
\includegraphics[width=1\columnwidth]{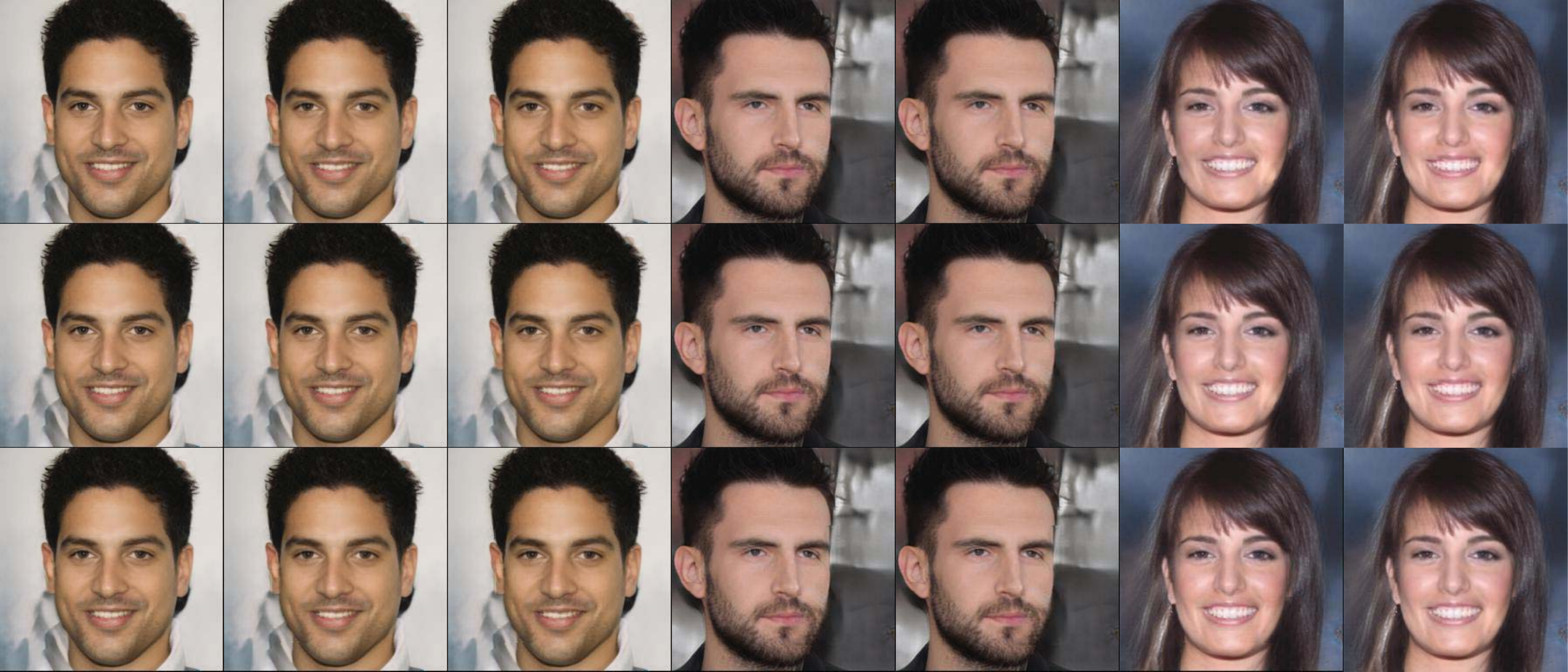}
\caption{Morphed facial images (IDs from the left to right in \textit{CelebA} are $011141$, $003668$, and $011910$, respectively.) with proprietary morphing fields measured by $\ell_2$-norm (values are $10, 120, 140, 170, 190, 300, 600$), $\ell_\infty$-norm (values are $0.2, 0.5, 1.0, 1.5, 2.0, 4.0, 6.0$), and multiplier $\delta$ (values are $0.2, 0.6, 0.8, 1.0, 1.2, 1.4, 2.0$), from top to bottom.}
\label{Figure:visual}
\end{figure}

\begin{figure}
\centering
\includegraphics[width=0.3\columnwidth]{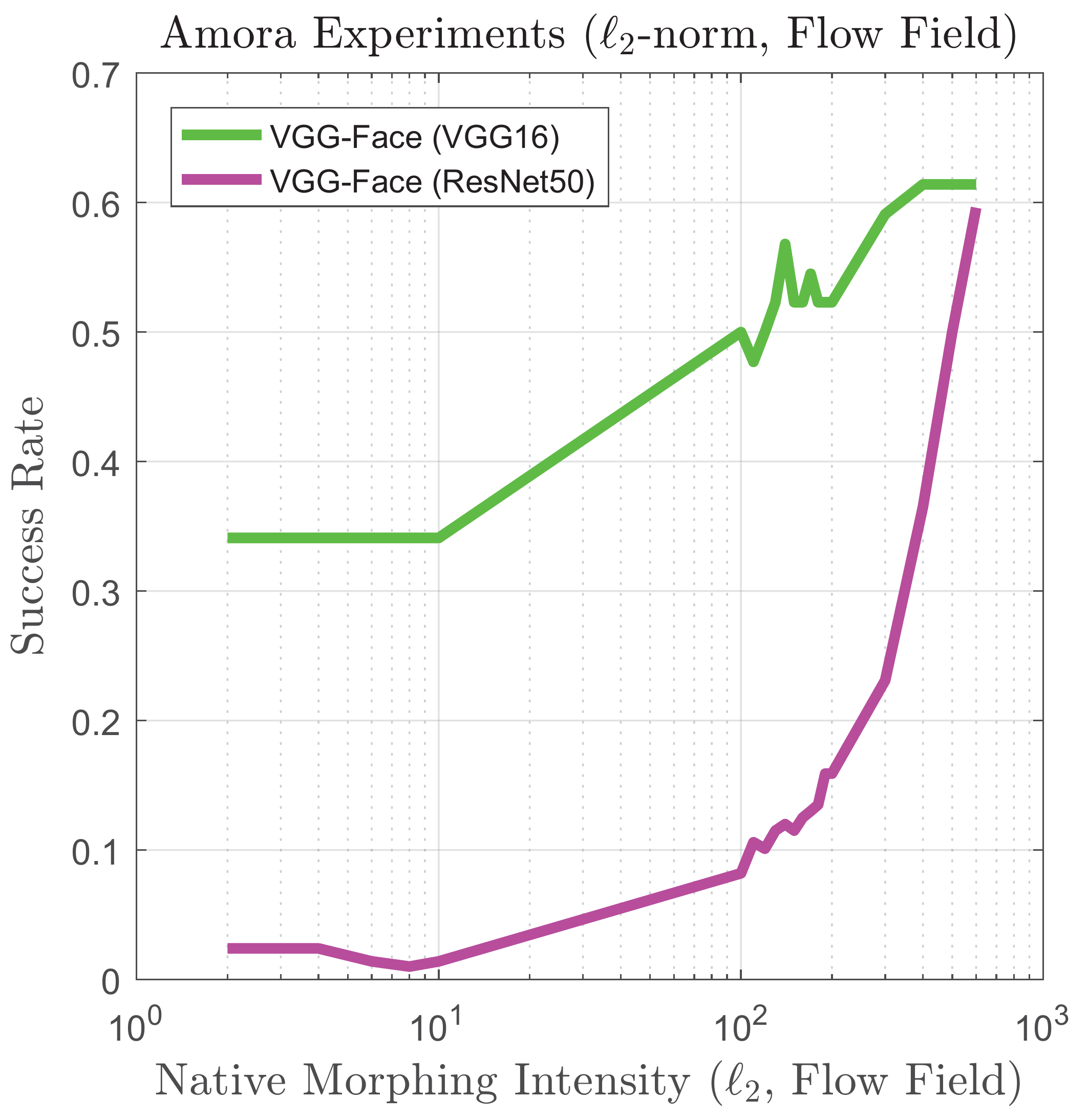}
\includegraphics[width=0.3\columnwidth]{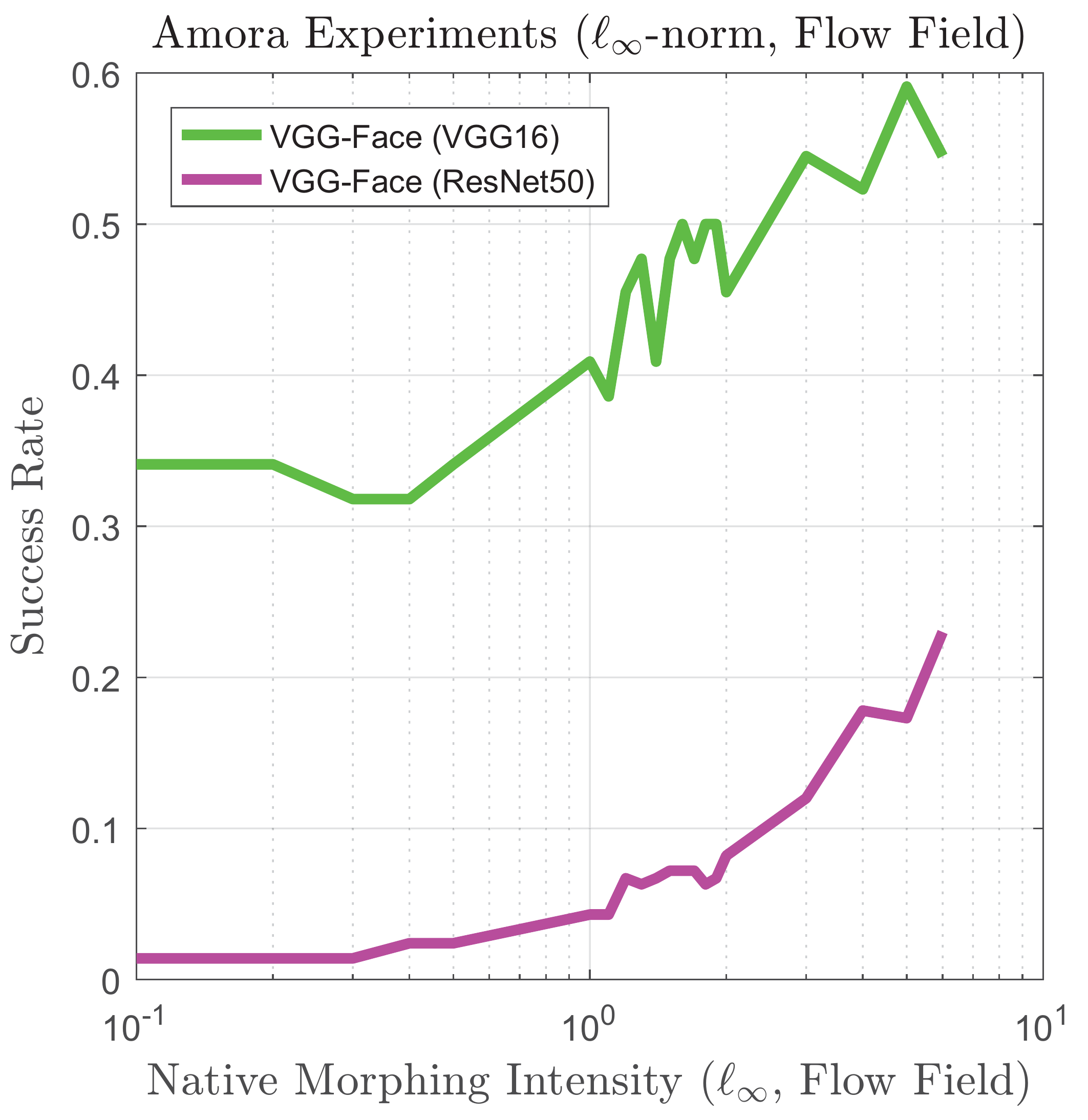}
\includegraphics[width=0.37\columnwidth]{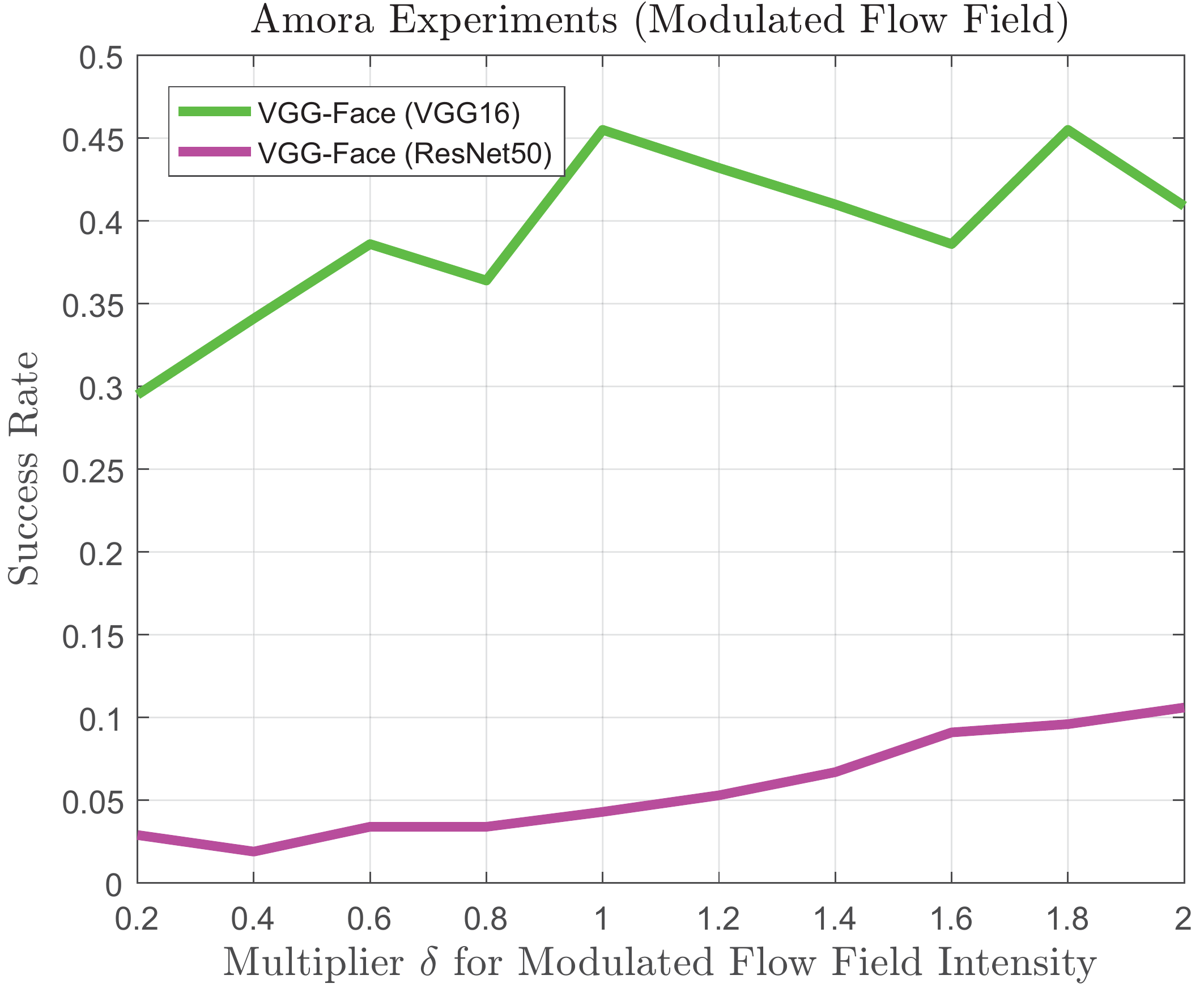}
\caption{The relation between the attack success rate and the modulated flow field on $\ell_2$-norm, $\ell_\infty$-norm, and multiplier $\delta$.}
\label{Figure:fig6}
\end{figure}

Table \ref{Table:relative_intensity} summarizes the results of attack success rate and the intensity of proprietary morphing with multiplier $\delta$. Figure~\ref{Figure:fig6} (R) plots a trend of the increase of the multiplier $\delta$ and attack success rate. We can also find that VGG-Face with VGG16 as backend architecture is much more vulnerable than VGG-Face with ResNet50 as backend architecture. To obtain an intuitive visualization of facial images morphed with proprietary morphing fields, we present some morphed facial images of three different identities with the three metrics in Figure~\ref{Figure:visual}.

\begin{table*}[t]
	\centering
    \small
	\caption{Attack success rate with different intensity on proprietary morphing field measured by $\ell_2$-norm.}
	\begin{tabular}{c|cccccccccccccccccccc}
	\toprule
		$\mathcal{M} \backslash \ell_2$ & 2 & 4 & 6 & 8 & 10 & 100 & 110 & 120 & 130 & 140 & 150 & 160 & 170 & 180 & 190 & 200 & 300 & 400 & 500 & 600  \\
	\midrule
		\begin{tabular}[c]{@{}c@{}}VGG16\end{tabular}    &   0.34  &  0.34   &  0.34   & 0.34    & 0.34    & 0.50    &  0.48    &  0.50   &  0.52 & 0.57 & 0.52 & 0.52 & 0.55 & 0.52 & 0.52 & 0.52 & 0.59 & 0.61 & 0.61 & 0.61      \\ \midrule

		\begin{tabular}[c]{@{}c@{}}ResNet50\end{tabular} &   0.02  &  0.02   &  0.01   & 0.01    &  0.01   & 0.08    &  0.11    &  0.10     &    0.12 & 0.12 & 0.12 & 0.13 & 0.13 & 0.14 & 0.16 & 0.16 & 0.23 & 0.37 & 0.50 & 0.60  \\
	\bottomrule
	\end{tabular}
	\label{Table:l2_to_rate}
\end{table*}

\begin{table*}[t]
	\centering
    \small
	\caption{Attack success rate with different intensity on proprietary morphing field measured by $\ell_\infty$-norm.}
	\begin{tabular}{c|cccccccccccccccccccc}
	\toprule
		$\mathcal{M} \backslash \ell_\infty$ & 0.1 & 0.2 & 0.3 & 0.4 & 0.5 & 1.0 & 1.1 & 1.2 & 1.3 & 1.4 & 1.5 & 1.6 & 1.7 & 1.8 & 1.9 & 2.0 & 3.0 & 4.0 & 5.0 & 6.0 \\
	\midrule
		\begin{tabular}[c]{@{}c@{}} VGG16\end{tabular}    &   0.34  &  0.34   &  0.32   & 0.32    & 0.34    & 0.41    &  0.39    &  0.46   &  0.48 & 0.41 & 0.48 & 0.50 & 0.48 & 0.50 & 0.50 & 0.46 & 0.55 & 0.52 & 0.59 & 0.55     \\ \midrule
		\begin{tabular}[c]{@{}c@{}} ResNet50\end{tabular} &   0.01  &  0.01   &  0.01   & 0.02    &  0.02   & 0.04 &0.04   &  0.07 & 0.06 & 0.07   &    0.07  & 0.07   & 0.07 & 0.06 & 0.07 & 0.08 & 0.12 & 0.18 & 0.17 & 0.23   \\
	\bottomrule
	\end{tabular}
	\label{Table:l_infinte_to_rate}
\end{table*}

\begin{table}[t]
	\centering
    \small
	\caption{Enhancing the intensity of proprietary morphing fields on the multiplier $\delta$.}
	\setlength{\tabcolsep}{2pt}
	\begin{tabular}{c|cccccccccc}
	\toprule
		$\mathcal{M} \backslash  \delta $ & 0.2 & 0.4 & 0.6 & 0.8 & 1.0 & 1.2 & 1.4 &1.6 & 1.8 & 2.0 \\
	\midrule
		\begin{tabular}[c]{@{}c@{}}VGG16\end{tabular}    &   0.295  &  0.341   &  0.386   & 0.364    & 0.455    & 0.432    &  0.410 & 0.386 & 0.455 & 0.409    \\ \midrule
		\begin{tabular}[c]{@{}c@{}}ResNet50\end{tabular} &  0.029  &  0.019   &  0.034   & 0.034    &  0.043   & 0.053 & 0.067 & 0.091  & 0.096  & 0.106 \\
	\bottomrule
	\end{tabular}
	\label{Table:relative_intensity}
\end{table}

\subsection{Evaluation of Transferabilities}

In this section, we discuss the transferabilities of our adversarial morphing attack. Transferability is an important property in adversarial attack, which is indicated by the degree that a successful attack in one model could be transferred to another model. In our experiment, we have demonstrated the effectiveness of our adversarial morphing attack by investigating the attack transferabilities between VGG-Face with VGG16 as backend architecture and VGG-Face with ResNet50 as backend architecture.
\begin{table}[t]
	\centering
    \small
	\caption{The transferabilities of our proposed adversarial morphing attack between VGG16 and ResNet50 with $\ell_2$-norm and $\ell_\infty$-norm.}
	\begin{tabular}{c|cc}
	\toprule
		\textit{metrics}$\backslash \mathcal{M}$ & \begin{tabular}[c]{@{}c@{}}VGG16\end{tabular}  &\begin{tabular}[c]{@{}c@{}}ResNet50\end{tabular}  \\
	\midrule
		$\ell_2$-norm      &  0.935  & 0.926   \\ \midrule
		$\ell_\infty$-norm &  0.893  & 0.885   \\
	\bottomrule
	\end{tabular}
	\label{Table:eval_transferbility}
\end{table}
Our transferability evaluation experiments are conducted on a dataset from Table \ref{Table:eval}. Each facial image is morphed with their proprietary morphing field with $\ell_2$-norm and $\ell_\infty$-norm as metrics to control the intensity of morphing field. Table \ref{Table:eval_transferbility} presents the experimental results in evaluating the transferabilities of adversarial morphing attack. The intensity of the proprietary morphing field is measured by $\ell_2$-norm and $\ell_\infty$-norm and their value are presented in Table \ref{Table:l2_to_rate} and Table \ref{Table:l_infinte_to_rate}, respectively. Experimental results show that our adversarial morphing attack achieves $90\%$ success rate at average, in attacking transferabilities evaluation.

\subsection{Open-set Evaluation}


In this section, we present the experimental results of our adversarial morphing attack in dealing with open-set attack where the focus is on unseen classes. We trained two popular FR systems with $500$ identities on \textit{CelebA-HQ} dataset, namely VGG-Face with VGG16 and ResNet50 as backend. In testing, the identity of a new given face is unseen in training. In the experiment, we evaluate whether our morphed facial images with assigned proprietary morphing fields decrease the performance of open-set FR systems in classification.
We use the receiver operating characteristic (ROC) curve to evaluate the performance of our Amora in attacking open-set FR systems. ROC curve is an important and common method for evaluating the performance of classifiers. Verification rate (VR) at 0.001 false accept rate (FAR), equal error rate (EER), and area under the ROC curve (AUC) are adopted as verification scores for evaluating the performance of our adversarial morphing attack in attacking open-set FR systems. Higher EER means higher attack success rate while lower VR and AUC mean higher attack success rate.
\begin{figure}[ht]
\centering
\includegraphics[width=0.48\columnwidth]{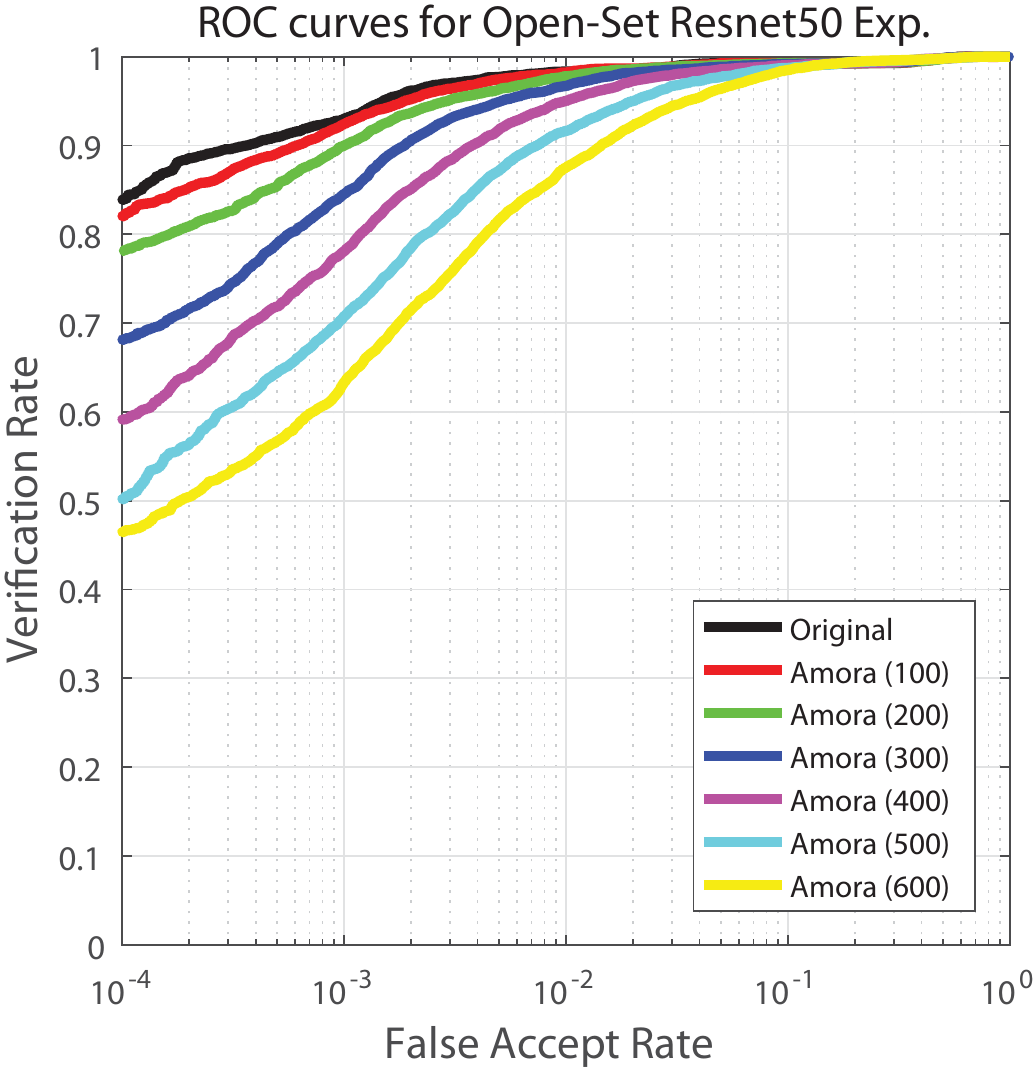}
\includegraphics[width=0.48\columnwidth]{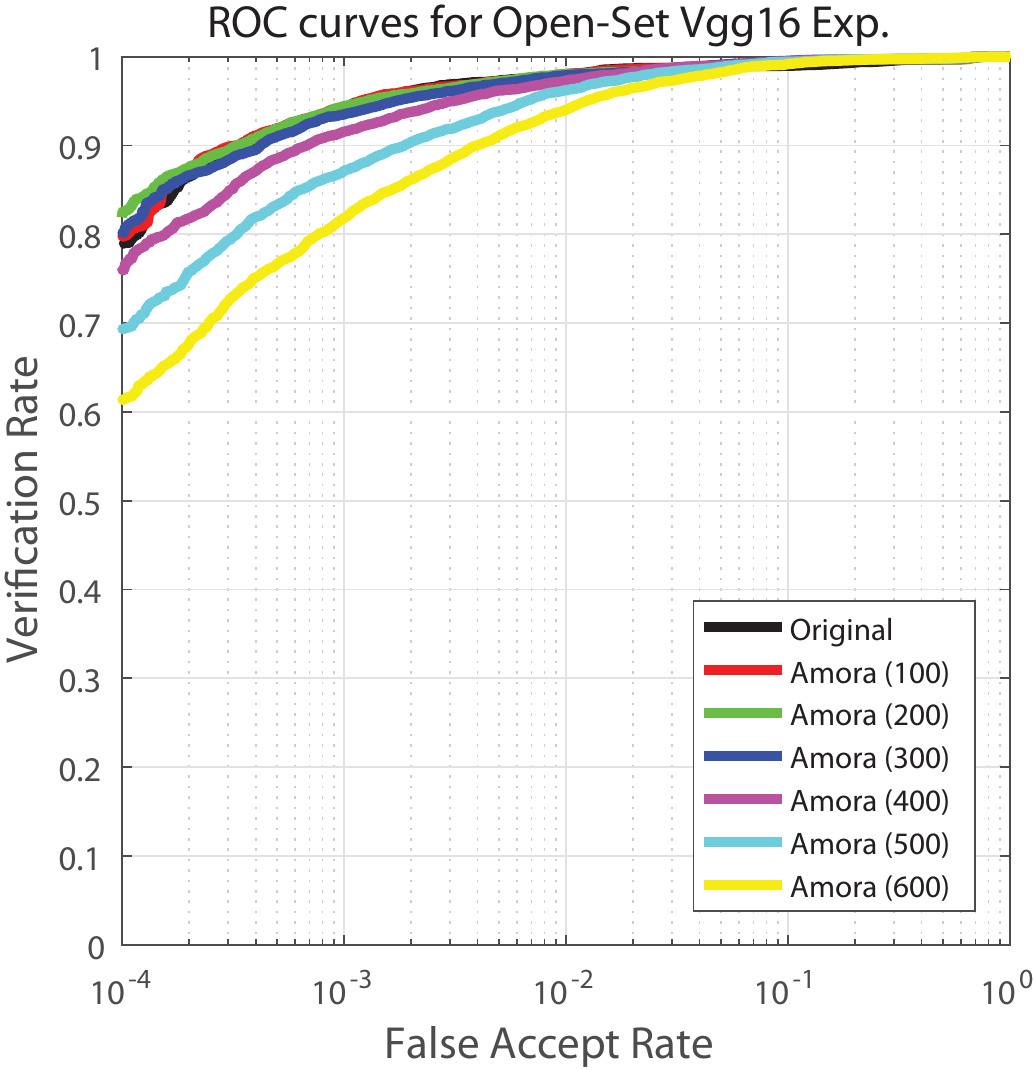}
\caption{ROC curves for the open-set experiments on the ResNet50 and VGG16 FR systems.}
\label{Figure:roc_resnet_vgg}
\end{figure}

Table \ref{Table:open_set_resnet_vgg} presents the open-set face verification scores with $20$ different intensities on proprietary morphing field measured by $\ell_2$-norm on ResNet50 and VGG16 face recognition systems, respectively. The two popular face recognition systems are more easily attacked when the intensity of proprietary morphing fields increased according to the face verification scores VR, EER, and AUC.
Figure~\ref{Figure:roc_resnet_vgg} shows the ROC curves for the open-set experiments on the ResNet50 and VGG16 FR systems, respectively. In Figure~\ref{Figure:roc_resnet_vgg}, we select six different intensities on the proprietary morphing field measured by $\ell_2$-norm to generate morphed facial images for attacking ResNet50 and VGG16 FR systems. The lower AUC means that the face recognition system is more easily attacked. In comparing with the original facial images, we find that our morphed facial images on different intensity proprietary morphing fields can achieve higher possibility in attacking open-set FR systems.
\begin{table*}[t]
	\centering
    \small
	\setlength{\tabcolsep}{2.8pt}
		\caption{Open-set face verification scores (verification rate (VR) at 0.001 false accept rate (FAR), equal error rate (EER), and area under the ROC curves (AUC)) with different intensity on proprietary morphing field measured by $\ell_2$-norm on ResNet50 and VGG16 FR systems, respectively.}
	\begin{tabular}{c|c|cccccccccccccccccccc|c}
	\toprule
	FR &  $\ell_2$ & 2 & 4 & 6 & 8 & 10 & 100 & 110 & 120 & 130 & 140 & 150 & 160 & 170 & 180 & 190 & 200 & 300 & 400 & 500 & 600 & Orig. \\
	\midrule
	\multirow{3}{*}{ \rotatebox[origin=c]{90}{ResNet50}} & VR  &   93.37  &  93.33   &  93.31   & 93.28    & 93.26    & 92.88    &  92.82   &  92.53   &  92.61 & 92.44 & 92.28 & 92.13 & 91.50 & 91.44 & 91.32 & 90.43 & 85.19 & 79.01 & 71.73 & 64.49  & 93.38   \\ \cmidrule{2-23}
	& EER    &   1.52  &  1.52   &  1.52   & 1.52    & 1.52    & 1.31    &  1.30  &  1.36  &  1.35 & 1.42 & 1.42 & 1.49 & 1.47 & 1.45 & 1.49 & 1.52 & 1.79 & 2.33 & 3.13 & 4.21 &  1.52    \\ \cmidrule{2-23}
	& AUC &   99.64  &  99.64   &  99.64   & 99.64    &  99.64   & 99.63 & 99.63 & 99.64   &  99.63 & 99.63 & 99.62   &    99.62  & 99.62   & 99.62 & 99.61 & 99.61 & 99.58 & 99.50 & 99.36 & 99.18  &  99.64  \\ \midrule \midrule
	\multirow{3}{*}{ \rotatebox[origin=c]{90}{VGG16}}& VR  &   94.33  &  94.34   &  94.40   & 94.38    & 94.36    & 94.71    &  94.74    &  94.76   &  94.80 & 94.78 & 94.83 & 94.80 & 94.73 & 94.73 & 94.64 & 94.66 & 93.71 & 91.83 & 87.52 & 82.66  & 94.69   \\ \cmidrule{2-23}
	& EER  &   1.48  &  1.46   &  1.47   & 1.44    & 1.44    & 1.48    &  1.49  &  1.55  &  1.54 & 1.54 & 1.58 & 1.59 & 1.60 & 1.59 & 1.63 & 1.66 & 1.70 & 1.80 & 2.13 & 2.73 &  1.42    \\ \cmidrule{2-23}
	& AUC &   99.64  &  99.64   &  99.64   & 99.64    &  99.65   & 99.68 & 99.68 & 99.69   &  99.69 & 99.69 & 99.69   &    99.69  & 99.69   & 99.69 & 99.69 & 99.69 & 99.71 & 99.70 & 99.66 & 99.56  &  99.62  \\
	\bottomrule
	\end{tabular}
	\label{Table:open_set_resnet_vgg}
\end{table*}

\subsection{Compared with Competitive Baselines}

In this section, we build two competitive baselines, permutation baseline and random baseline, to investigate whether our proprietary morphing fields can indeed be effectively served as guidance in morphing facial images. In the experiments, we mainly 1) explore the attack success rate and the intensity of morphing fields measured by $\ell_2$-norm and $\ell_\infty$-norm, 2) investigate the attack success rate and the visual quality of morphed facial images with structural similarity index (SSIM) and normalized cosine similarity (NCS). Here, we consider a \emph{region of operation} (ROO) in evaluating the performance of our proprietary morphing field compared with baselines. ROO is the region where `adversarial attack' assumption holds, \ie, imperceptible to human eyes.

\begin{figure}
\centering
\includegraphics[width=0.48\columnwidth]{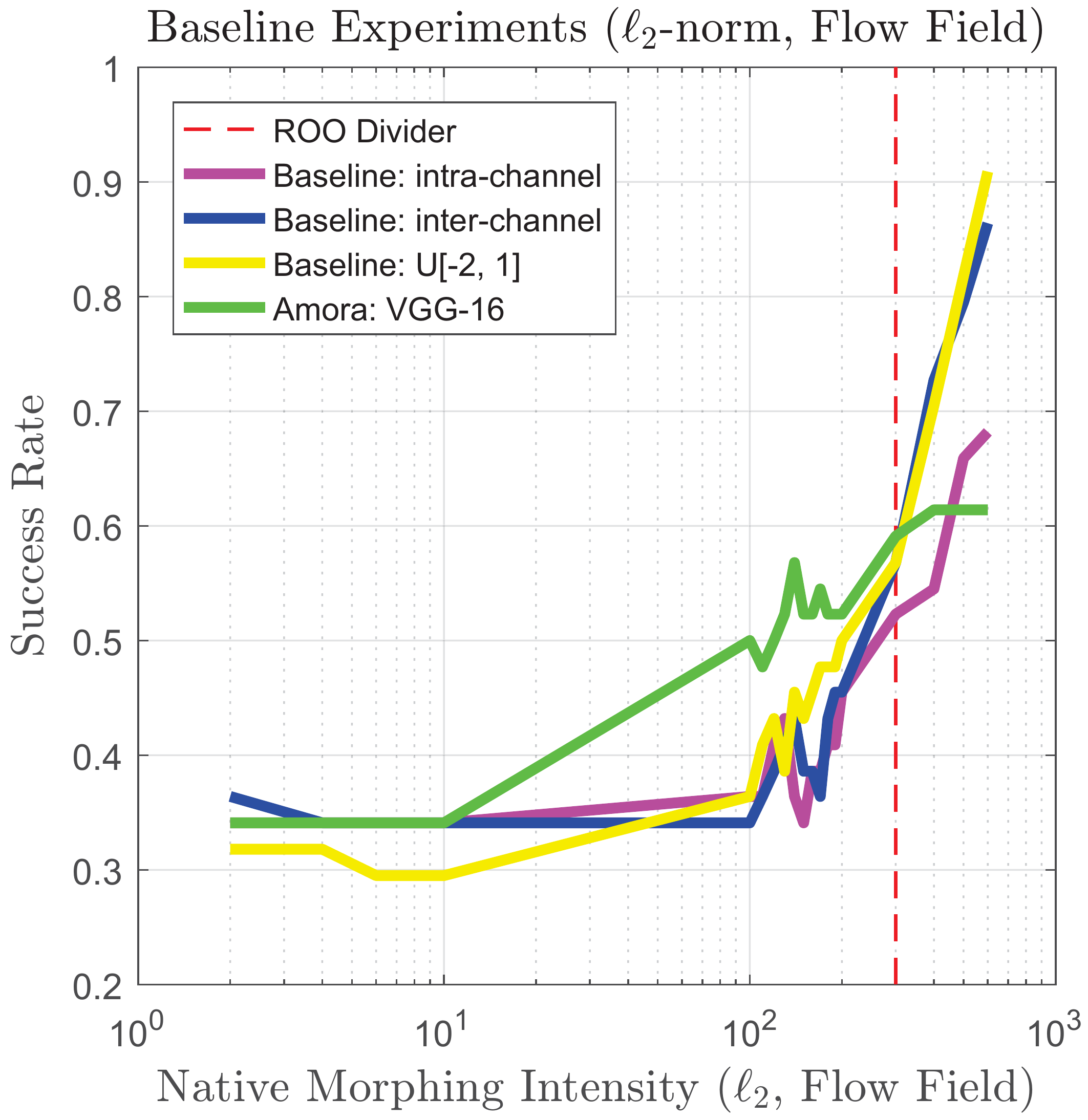}
\includegraphics[width=0.48\columnwidth]{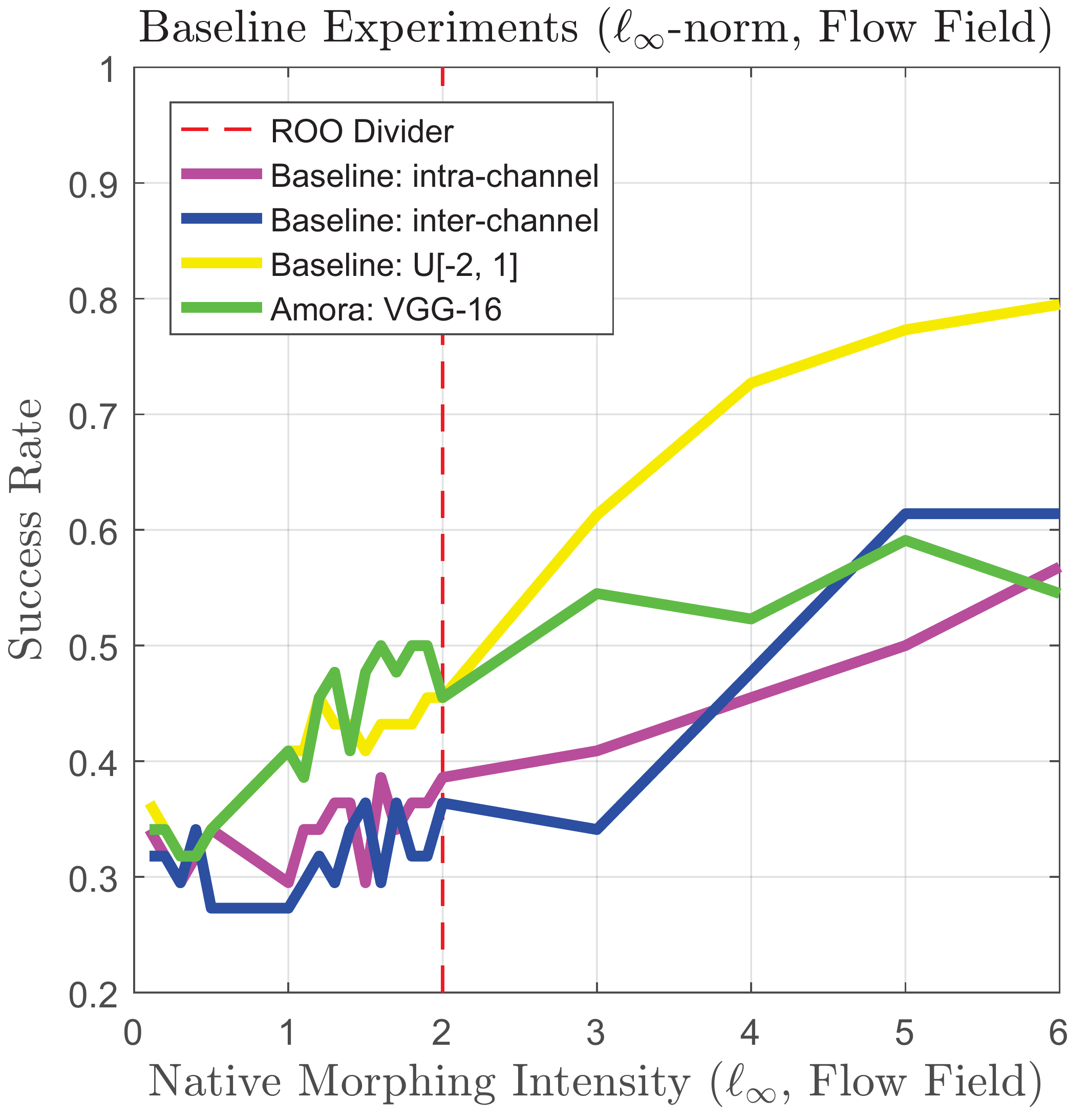}
\caption{Baseline experiment results on $\ell_2$-norm and $\ell_\infty$-norm in attacking VGG16 FR system. Baseline: intra-channel and inter-channel are permutation baseline, Baseline: U[-2,1] is random baseline. Left side of the ROO Divider is the region of operation where adversarial assumption holds. Amora trumps in the ROO.}
\label{Figure:baseline}
\vspace{-10pt}
\end{figure}

Permutation baseline permutes the proprietary morphing fields while maintaining their intensity the same as the original proprietary morphing fields. As the proprietary morphing field includes horizontal channel $\mathbf{f}^h$ and vertical channel $\mathbf{f}^v$, permutation baseline can permute morphing fields within channels, named \emph{intra-channel} permutation baseline, and between two channels, named \emph{inter-channel} permutation baseline. Random baseline randomly generated morphing field $\mathbf{f}^h$ and $\mathbf{f}^v$, which follow a uniform distribution $\mathcal{U}[-2,1]$ as more than 94.4\% raw morphing fields lying in this range in the query stage. Figure~\ref{Figure:baseline} presents the results of our proprietary morphing field and two baselines in attacking the VGG16 FR system.

Experimental results in Figure~\ref{Figure:baseline} demonstrate that our proprietary morphing field performs better than the permutation and random baseline in ROO (see the ROO divider). Being able to outperform the baselines within the region of operation is what truly matters.
Figure~\ref{Figure:baseline} also illustrates that the two chosen baselines also show its power in attacking FR systems. However, the two chosen baselines are considered strong baselines as they capitalize on the prior knowledge of what optical flow fields should be. It is interesting to explore some moderate baselines in our future work. Table \ref{Table:baseline_ssim_ncs_success_rate} presents the experimental result in comparison with baselines where the similarity distance between the original facial image and morphed facial images is measured with SSIM and NCS. It indicates that our proprietary morphing fields achieve a high attack success rate when the morphed facial images are similar to original facial images, especially the SSIM and NCS values are larger than 0.9 and 0.998, respectively. Thus, our proprietary morphing field outperforms the three random baselines when the morphed facial images appear subtle change.
\begin{table}[t]
	\centering
    \small
	\setlength{\tabcolsep}{0.6pt}
	\caption{Attack success rate and the visual quality of morphed facial images measured with SSIM and NCS with three random baselines, $\mathcal{U}[-2,1]$, $\mathcal{U}[-1,1]$, and $\mathcal{N}(0,1)$. Large SSIM and NCS values denote that the morphed facial image is similar to the original face. \textcolor{dg}{Green} region is the ROO.}
	\begin{tabular}{c|ccccc}
	\toprule
		$\mathcal{M} \backslash$ssim & \textcolor{red}{[0.75, 0.8)} & \textcolor{red}{[0.8, 0.85)} & \textcolor{red}{[0.85, 0.9)} & \textcolor{dg}{[0.9, 0.95) }& \textcolor{dg}{[0.95, 1.0]} \\ \midrule
		\begin{tabular}[c]{@{}c@{}} VGG-16\end{tabular}    &  0 & 0 & 0.048 & 0.073 & 0.373  \\ \midrule
		\begin{tabular}[c]{@{}c@{}} $\mathcal{U}$[-2,1]\end{tabular}    &   0.278 & 0.128 & 0.079 & 0.032 & 0.046 \\ \midrule
		\begin{tabular}[c]{@{}c@{}} $\mathcal{U}$[-1,1]\end{tabular}    &   0.346 & 0.181 & 0.104 & 0.073 & 0.035 \\ \midrule
		\begin{tabular}[c]{@{}c@{}} $\mathcal{N}$(0,1)\end{tabular}    &   0.224 & 0.123 & 0.07 & 0.016 & 0.032 \\ \midrule \midrule
		$\mathcal{M} \backslash$ncs &   \textcolor{red}{[0.990, 0.992)} & \textcolor{red}{[0.992, 0.994)} & \textcolor{red}{[0.994, 0.996)} & \textcolor{red}{[0.996, 0.998)} & \textcolor{dg}{[0.998, 1.0]}\\ \midrule
		\begin{tabular}[c]{@{}c@{}} VGG-16\end{tabular}    &   0 & 0.001 & 0.014 & 0.058 & 0.420  \\ \midrule
		\begin{tabular}[c]{@{}c@{}} $\mathcal{U}$[-2,1]\end{tabular}    &   0.032 & 0.091 & 0.185 & 0.355 & 0.222  \\ \midrule
		\begin{tabular}[c]{@{}c@{}} $\mathcal{U}$[-1,1]\end{tabular}    &  0 & 0.006 & 0.046 & 0.279 & 0.444  \\ \midrule
		\begin{tabular}[c]{@{}c@{}} $\mathcal{N}$(0,1)\end{tabular}    &   0.008 & 0.052 & 0.151 & 0.366 & 0.312 \\
	\bottomrule
	\end{tabular}
	\label{Table:baseline_ssim_ncs_success_rate}
	\vspace{-10pt}
\end{table}

\section{Conclusions}\label{sec:concl}

In this work, we introduced and investigated a new type of black-box adversarial attack to evade deep-learning based FR systems by morphing facial images with learned optical flows. The proposed attack morphs/deforms pixels spatially as opposed to adversarial noise attack that perturbs the pixel intensities.
With a simple yet effective joint dictionary learning pipeline, we are able to obtain a proprietary morphing field for each individual attack. Experimental results have shown that some popular FR systems can be evaded with high probability and the performance of these systems is significantly decreased with our attacks. Our observation raises essential security concerns in the current FR systems. Through comprehensive evaluations, we show that a black-box adversarial morphing attack is not only possible, but also compromises the FR systems significantly.

The proposed black-box adversarial morphing attack points to an orthogonal direction that can complement the existing adversarial noise attacks as well as other adversaries such as DeepFakes \cite{fakepolisher,deeprhythm} and novel non-additive attacks \cite{guo2020abba,pasadena,guo2019spatial}. Therefore, it is possible to combine various attack types in the future. Furthermore, how existing DL gauging tools \cite{ma2018deepgauge,xie2019deephunter,ma2018deepmutation,deepct} can help further improve the proposed attack modality is also worth studying.

While presentation spoofing attacks are relatively easier to be defended because they heavily rely on physical props, adversarial noise attacks are less likely to be presented in real-world setups. Therefore, it is useful to perform more extensive studies of physical attacks based on adversarial morphing, which performs semantically coherent attack with local facial deformation and is likely to occur in real scenarios such as an expression filter to attack mobile face authentication application (mobile payment/ social media), in tandem with freeform optics that bends light \cite{brand2019freeform}, \etc.

\begin{acks}
This research was supported in part by Singapore National Cybersecurity R\&D Program No. NRF2018NCR-NCR005-0001, National Satellite of Excellence in Trustworthy Software System No. NRF2018NCR-NSOE003-0001, NRF Investigatorship No. NRFI06-2020-0022. It was also supported by JSPS KAKENHI Grant No. 20H04168, 19K24348, 19H04086, and JST-Mirai Program Grant No. JPMJMI18BB, Japan. We gratefully acknowledge the support of NVIDIA AI Tech Center (NVAITC) to our research.
\end{acks}


\bibliographystyle{ACM-Reference-Format}
\bibliography{ref_amora}


\end{document}